\pdfoutput=1
\documentclass{article}
\usepackage{ijcai24}

\usepackage{times}
\usepackage{soul}
\usepackage{url}
\usepackage[hidelinks]{hyperref}
\usepackage[utf8]{inputenc}
\usepackage[small]{caption}
\usepackage{graphicx}
\usepackage{amsmath}
\usepackage{amsthm}
\usepackage{booktabs}
\usepackage{algorithm}
\usepackage{algorithmic}
\usepackage[switch]{lineno}

\usepackage{natbib}
\usepackage{arydshln}
\usepackage{newfloat}
\usepackage{listings}
\usepackage{mathtools} 
\usepackage{adjustbox} 
\usepackage{multirow} 
\usepackage{dsfont}
\usepackage{tikz}
\usepackage{pgfplots}
\usepackage{booktabs} 
\usepackage{xcolor} 
\usepackage[most]{tcolorbox}
\definecolor{blue1}{rgb}{0.15,0.15,0.15}
\definecolor{blue2}{rgb}{0.30,0.30,0.30}
\definecolor{blue3}{rgb}{0.45,0.45,0.45}
\definecolor{blue4}{rgb}{0.60,0.60,0.60}
\definecolor{blue5}{rgb}{0.70,0.70,0.70}
\definecolor{blue6}{rgb}{0.80,0.80,0.80}
\definecolor{egggreen}{HTML}{d7e6e8}
\definecolor{lightskyblue3}{HTML}{dbdbeb}
\definecolor{lightyellowgreen3}{HTML}{d6ecf0}
\definecolor{constraint}{HTML}{eee2f0}
\definecolor{LemonChiffon2}{HTML}{EEE9BF}
\definecolor{Ivory2}{HTML}{EEEEE0}
\definecolor{LavenderBlush2}{HTML}{EEE0E5}
\definecolor{RED}{HTML}{FCBBC3}
\newcommand\InputName[1]{\colorbox{Ivory2}{#1}}
\newcommand\LabelSet[1]{\colorbox{LemonChiffon2}{#1}}
\newcommand\TaskA[1]{\colorbox{LavenderBlush2}{#1}}
\newcommand\TaskB[1]{\colorbox{lightyellowgreen3}{#1}}
\newcommand\Regulation[1]{\colorbox{egggreen}{#1}}
\newcommand\Answer[1]{\colorbox{lightskyblue3}{#1}}
\newcommand\LabelConstraint[1]{\colorbox{constraint}{#1}}

\newtcolorbox{mybox}[1][]{
	width=\columnwidth,
	colback = gray!6, 
	colframe = black, 
	boxrule = 0.8pt,
	boxsep=0pt,left=10pt,right=10pt,top=8pt,bottom=8pt,
	fontupper=\linespread{1.2}\selectfont,
	title=#1}

\usepackage{tikz}
\usepackage[edges]{forest} 

\usepackage{pgfplots}

\usepackage{makecell}

\usepackage[most]{tcolorbox}
\definecolor{blue1}{rgb}{0.15,0.15,0.15}
\definecolor{blue2}{rgb}{0.30,0.30,0.30}
\definecolor{blue3}{rgb}{0.45,0.45,0.45}
\definecolor{blue4}{rgb}{0.60,0.60,0.60}
\definecolor{blue5}{rgb}{0.70,0.70,0.70}
\definecolor{blue6}{rgb}{0.80,0.80,0.80}
\definecolor{red}{rgb}{1.00, 0.00, 0.00}
\definecolor{green}{rgb}{0.00, 0.45, 0.00}
\definecolor{LightPurple}{rgb}{0.75, 0.33, 0.82}
\newtcolorbox{taskbox}[2][]{%
	enhanced, breakable,
	colframe=gray!25,
	colback=gray!2,
	arc=1mm,
	outer arc=1mm,
	fontupper=\linespread{1.1}\selectfont,
	fontlower=\small,
	coltitle=blue1,
	fonttitle=\bfseries,
	boxsep=2mm,
	left=2mm,
	right=2mm,
	top=0mm,
	bottom=0mm,
	before={\noindent},
	segmentation style={solid, blue3},
	title=#2,%
	#1
}

\title{CroPrompt: Cross-task Interactive Prompting for Zero-shot \\ Spoken Language Understanding}

\author{
Libo Qin$^{\clubsuit}$\thanks{\ \ Equal Contribution}
\and
Fuxuan Wei$^{\spadesuit}$\footnotemark[1]
\and
Qiguang Chen$^{\spadesuit}$\and
Jingxuan Zhou$^{\clubsuit}$\and
Shijue Huang$^{\diamondsuit}$\and
Jiasheng Si$^{\heartsuit}$ \and 
Wenpeng Lu$^{\heartsuit}$ \and
Wanxiang Che$^{\spadesuit}$
\\
\affiliations
$^{\clubsuit}$School of Computer Science and Engineering, Central South University, China\\
$^{\spadesuit}$Research Center for Social Computing and Information Retrieval, Harbin Institute of Technology, China\\
$^{\diamondsuit}$Harbin Institute of Technology, Shenzhen, China\\
$^{\heartsuit}$ Key Laboratory of Computing Power Network and Information Security, Ministry of Education, Qilu University of Technology (Shandong Academy of Sciences)
\emails
lbqin@csu.edu.cn,
fxwei@ir.hit.edu.cn
}

\begin{document}

\maketitle

\begin{abstract}
	Slot filling and intent detection are two highly correlated tasks in spoken language understanding (SLU).	
	Recent SLU research attempts to explore zero-shot prompting techniques in large language models to alleviate the data scarcity problem.
	Nevertheless, the existing prompting work ignores the cross-task interaction information for SLU, which leads to sub-optimal performance.
	To solve this problem, we present the pioneering work of Cross-task Interactive Prompting (CroPrompt) for SLU, which enables the model to interactively leverage the information exchange across the correlated tasks in SLU. 
	Additionally, we further introduce a multi-task self-consistency mechanism to mitigate the error propagation caused by the intent information injection.
	We conduct extensive experiments on the standard SLU benchmark and the results reveal that CroPrompt consistently outperforms the existing prompting approaches.
	In addition, the multi-task self-consistency mechanism can effectively ease the error propagation issue, thereby enhancing the performance.
	We hope this work can inspire more research on cross-task prompting for SLU.
\end{abstract}
\section{Introduction}
Intent detection and slot filling are two related tasks in spoken language understanding (SLU), which are used to extract the slots and intent of users' utterance to help the dialogue system to generate correct system response~\citep{ijcai2021p622}.
Take the input utterance ``\textit{Find the movie The Ghost}'', the former task can be viewed as  a sentence classification to capture the intents (e.g., \texttt{SearchMovie}) of the user while the latter can be modeled as a sequence labeling task to extract a set of slots (e.g, \texttt{movie\_name=The Ghost}).

\begin{figure}[t]
	\centering
	\includegraphics[width=0.42\textwidth]{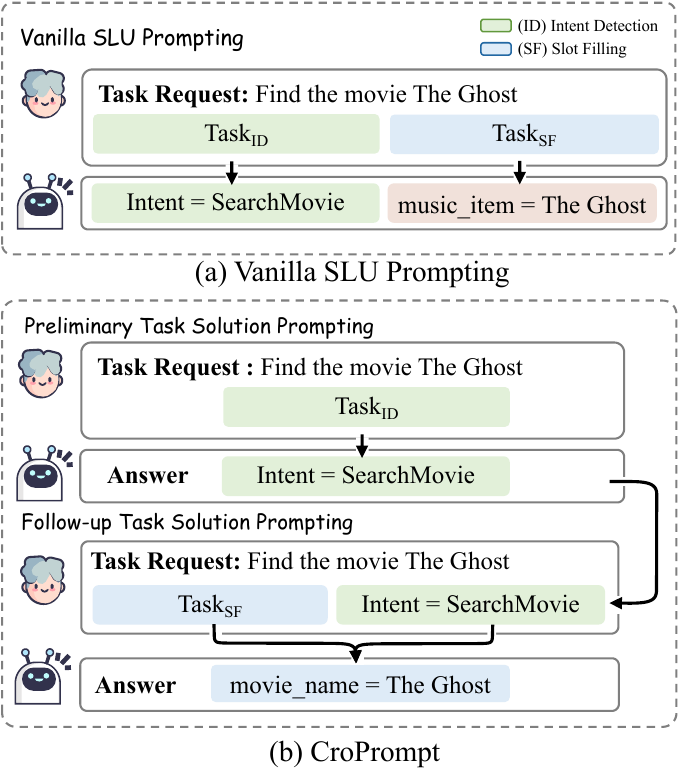}
	\caption{(a) Vanilla SLU Prompting directly utilizes a single conversation turn for prompting intent detection and slot filling without any interaction information while (b) CroPrompt 
		considers explicit interaction across multiple tasks by directly incorporating the result of intent detection for slot filling. 
	}
	\label{fig:intro}
\end{figure}

Since the two tasks are highly tied, prevailing SLU approaches in the literature mainly consider the relationship between joint slot filling and intent detection~\citep{goo-etal-2018-slot,DBLP:conf/naacl/WangSJ18,DBLP:conf/emnlp/LiLQ18,DBLP:conf/acl/ENCS19,DBLP:conf/emnlp/LiuMZZCX19}. 
The existing joint models can be categorized into two main categories. The first category comprises a series of models that model implicit interaction for SLU by simply considering a vanilla multi-task paradigm for intent detection and slot filling without any explicit interaction module~\citep{DBLP:conf/ijcai/ZhangW16a,DBLP:journals/corr/LiuL16d,liu-lane-2016-joint,DBLP:conf/interspeech/Hakkani-TurTCCG16}. 
The second category includes models that introduce an explicit interaction module (e.g., intent$\to$slot or intent $ \leftrightarrow$ slot task interaction) to explicitly build information flow across intent detection and slot filling, achieving superior performance~\citep{goo-etal-2018-slot,9414110}.

Though promising performance has been achieved, existing SLU approaches still heavily rely on lots of annotated data for training, which is hard to collect. Recently,  Prompt-based methods relying on large language models (LLMs) have shown remarkable performance on zero-shot settings, which reduces the time and effort for data annotation.
Inspired by this, \citet{pan2023preliminary} introduce a vanilla SLU prompting to prompt the intent detection and slot filling simultaneously in a single turn, which is shown in Figure~\ref{fig:intro}(a).
While \citet{pan2023preliminary} take the first meaningful step towards zero-shot SLU with LLM, the vanilla SLU prompting approach still faces a major drawback: \textit{neglecting the explicit exchange of information among correlated tasks, thereby limiting their performance}. Intuitively, when two tasks are highly correlated, the information from one task can be utilized to enhance the performance of the other related task. 
Consequently, it is promising to explicitly leverage the interaction information of correlated tasks in prompting for SLU.

Motivated by the observation, in this work, we introduce a novel cross-task interactive prompting approach (\textbf{CroPrompt}) in SLU.
As shown in Figure~\ref{fig:intro} (b), in contrast to the vanilla SLU prompting technique that directly prompts LLM to generate results for all tasks simultaneously, CroPrompt is capable of incorporating interactive information exchange from intent detection to slot filling.
To be specific, initially, CroPrompt prompts the LLM to obtain the results of the initial task (intent detection), followed by subsequently generating outcomes for the related task (slot filling) conditioned on the outputs of the intent detection, which naturally leverages the information exchange across related tasks.
Since the explicit incorporation of intent information could potentially result in error propagation, inspired by self-consistency technique~\citep{wang2022self}, we further introduce a multi-task consistency prompting to integrate diverse pathways of reasoning to address the error propagation issue.

We evaluate CroPrompt on the standard SLU benchmark.
The experimental results demonstrate that CroPrompt consistently outperforms previous prompting methods. In addition, when combined with multi-task self-consistency, the performance is further improved.

Contribution of this work can be summarized as:
\begin{itemize}
	\item [(1)] This work 
	represents a pioneering effort in explicit cross-task prompting for zero-shot SLU and we introduce a cross-task interactive prompting (CroPrompt) to this end;
	\item [(2)] In contrast to the previous studies, CroPrompt has the advantage of  interactively leveraging the information exchange across related tasks, thereby enhancing performance for slot filling and intent detection;
	\item [(3)] Furthermore, we introduce a multi-task consistency strategy to alleviate the potential error propagation issue by the intent information injection;
	\item [(4)] Extensive experimental results show that CroPrompt can consistently outperform previous prompting methods and achieve superior performance. In addition, the integration of multi-task self-consistency further enhances the performance.

\end{itemize}

\begin{figure*}[htbp]
	\centering
	\includegraphics[width=.95\textwidth]{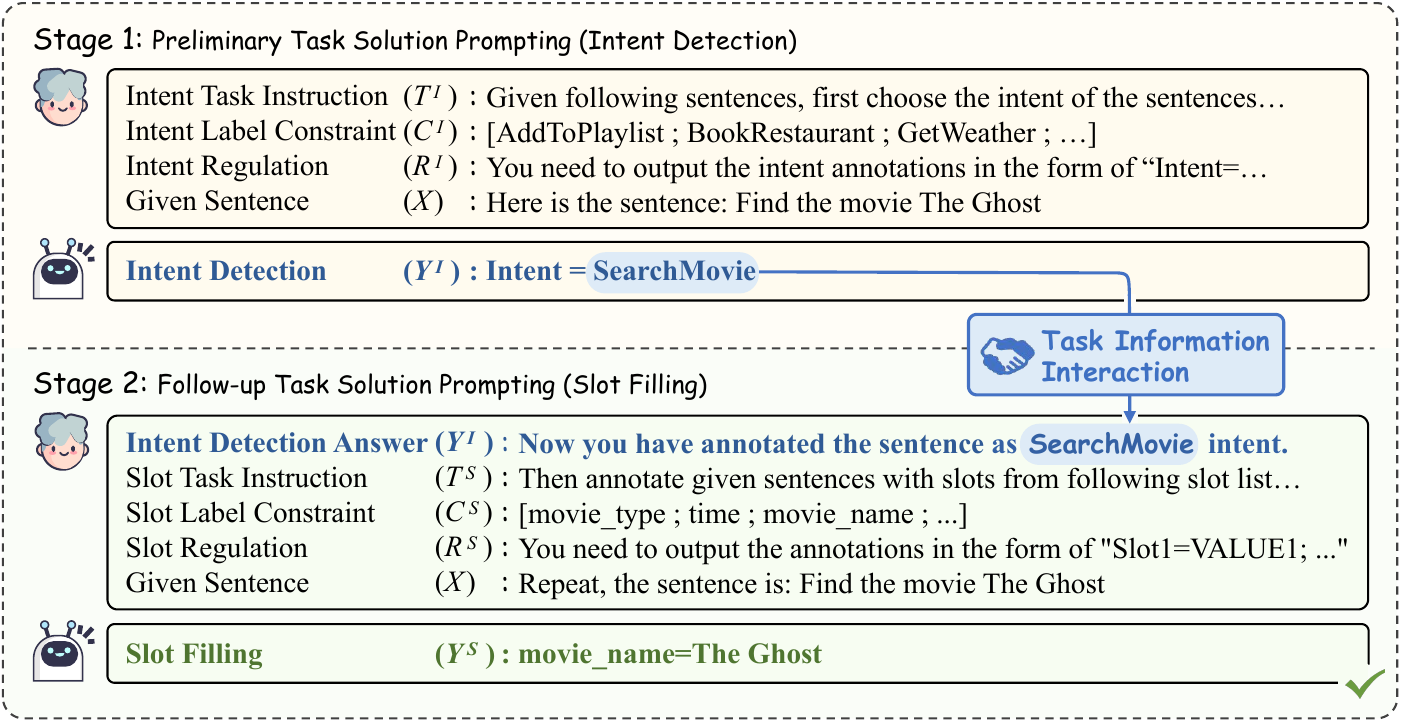}
	\caption{The workflow of CroPrompt, which first utilizes the \textit{preliminary task solution prompting} for intent detection and then the \textit{follow-up task solution prompting} is introduced for slot filling conditioned with the predicted intent results by task information interaction.}
	\label{fig:main}
\end{figure*}

\section{CroPrompt}
\label{section:mmprompt}
This section illustrates the workflow of CroPrompt, which consists of two stages: (1) \textit{stage 1: preliminary task solution prompting} ($\S \ref{sec:preliminary-task-solution-prompting}$) and (2) \textit{stage 2: follow-up task solution prompting} ($\S \ref{sec:follow-up-task-solution-prompting}$) to explicitly leverage the information exchange across related  tasks, which is shown in Figure~\ref{fig:main}.

\subsection{Stage 1: Preliminary Task Solution Prompting} \label{sec:preliminary-task-solution-prompting}
Since the slot filling is highly related to the intent detection, we first use \textit{preliminary task solution prompting} to obtain the result of intent detection, which can be further used for guiding slot filling.
Formally, the input format of \textit{preliminary task solution prompting} is shown as:
\begin{mybox}
	\TaskA{[Intent Task Instruction $T^I$]}: Given following sentences, first choose the intent of the sentences ..\\
	\LabelSet{[Intent Label Constraint $C^I$]}: [AddToPlaylist; ...]\\
	\Regulation{[Intent Regulation $R^I$]}:
	You need to output the intent annotations in the form of \Regulation{["Intent=..."]}\\
	\InputName{[Given Sentence $X$]}: Here is the sentence: ...
\end{mybox}
Each part of the prompt is introduced as follows:
\begin{itemize}
	\item [(1)] \TaskA{Intent Task Instruction $T^I$} describes the definition of intent detection, aiming to help LLM to clearly understand the intent detection task.
	\item [(2)] \LabelSet{Intent Label Constraint $C^I$} contains label set $L^I$ from the intent detection task.
	\item [(3)] \Regulation{Intent Regulation $R^I$} is provided to ensure the model generates standardized answers for the unified evaluation. 
	\item [(4)] \InputName{Given Sentence $X$} denotes the test input instance the model needs to address.
\end{itemize}

In summary, the formula for \textit{Preliminary Task Solution Prompting} can be expressed as follows:
\begin{equation}
	\mathcal{Y}^I = \arg\max_{L^I} p(l^I_i|T^I, C^I, R^I, X),
\end{equation}
where $\mathcal{Y}^I$ denotes the predicted intent with regulation $R$; ${l^I_i}$ denotes each intent label.

\subsection{Stage 2: Follow-up Task Solution Prompting}\label{sec:follow-up-task-solution-prompting}

After obtaining the intent prediction, we further introduce a \textit{follow-up task solution prompting} to explicitly inject intent information to assist slot filling.
Specifically, the  \textit{follow-up task solution prompting} can be defined as:
\begin{mybox}
	\Answer{[Intent detection Answer $\mathcal{Y}^I$]}:
	\ Now you have annotated ... as \Answer{[Answer $\mathcal{Y}^I$]}.\\
	\TaskB{[Slot Task Instruction $T^S$]}: Then annotate given sentences with slots from following slot list... \\
	\LabelConstraint{[Slot Label Constraint $C^S$]}: movie\_type:... \\
	\Regulation{[Slot Regulation $R^S$]}
	You need to output ... in the form of \Regulation{["Slot1=Value1;..."]}\\
	\InputName{[Given Sentence $X$]}:
	Repeat, the sentence is ... 
\end{mybox}
Similarly, the prompt is introduced in detail as follows:
\begin{itemize}
	\item [(1)] \Answer{Intent detection Answer $\mathcal{Y}^I$} denotes the predicted intent answer from the \textit{preliminary task solution prompting} step, which is the core contribution of CroPrompt to explicitly utilize the information of intent detection.
	\item [(2)] \TaskB{[Slot Task Instruction $T^S$]} describes the definition of slot filling task.
	\item [(3)] \LabelConstraint{Slot Label Constraints $C^S$} consists of the slot label set for model to choose. Notably, in contrast to the \LabelSet{Slot Label Constraints  $C^I$} in \textit{Preliminary Task Solution Prompting} step,  we only present a subset of labels $L^S$ that are related to the predicted \Answer{answer $\mathcal{Y}^I$}, rather than the entire list of slots, which can greatly reduce the search space.
	For example, if we first predict the intent as \texttt{SearchMovie} in \textit{Preliminary Task Solution Prompting} stage, we can only search for slots related to \texttt{SearchMovie}, rather than the entire set of slot labels in the dataset in the \textit{Follow-up Task Solution Prompting} stage, greatly reducing the search space.
	\item [(4)] \Regulation{Slot Regulation $R^S$} is used to enforce output formatting of slot filling.
	\item [(5)] \InputName{Given Sentence $X$} is the test instance of slot filling.
\end{itemize}

The formalization of slot filling prediction for Task 2 is described below:

\begin{small}
	\begin{equation}
		\mathcal{Y}^S\!\! =\! \mathop{argmax}_{L^S} p(y^S_{1}, \dots, y^S_{n}|\mathcal{Y}^I, T^S, C^S, R^S, X),
	\end{equation}
\end{small}
where $\mathcal{Y}^S$ denotes the slot filling result.

Compared to the previous method which accomplishes both intent recognition and slot filling tasks in a single conversation turn, our CroPrompt has the following advantages: (1) CroPrompt is capable of capturing task information interaction across intent detection and slot filling where the answer of intent detection provides additional valuable input information for slot filling (2) Simultaneously, the label space of slot filling task can be significantly reduced by our prompt architecture, leading to shorter prompt length and easier comprehension by the LLM.

\subsection{Multi-Task Self-Consistency Prompting}
\label{sec:consistent}

\begin{figure}[htbp]
	\centering
	\includegraphics[width=0.45\textwidth]{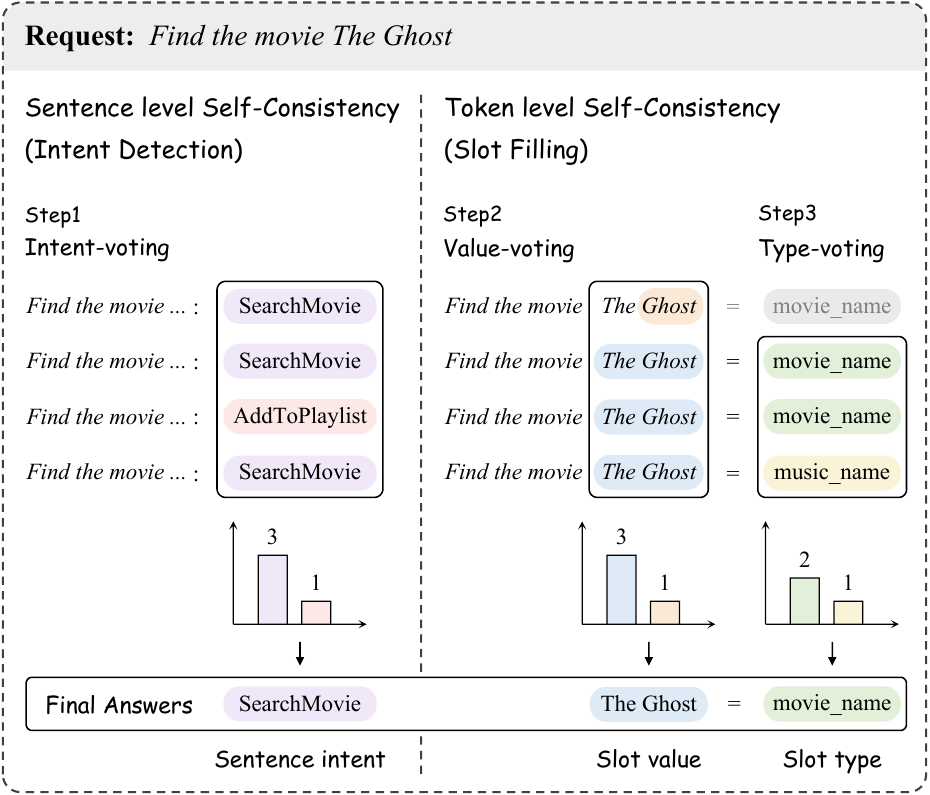}
	\caption{Multi-Task Self-Consistency Prompting for Intent Detection and Slot Filling.}
	\label{fig:consistency}
\end{figure}

Since CroPrompt explicitly introduces intent results for slot filling, it may
lead to error propagation if wrong intent information is introduced. Therefore, 
inspired by self-consistency~\citep{wang2023selfconsistency}, we propose multi-task consistency (MT-Self-Consistency) learning to integrate inference results from different pathways to mitigate the potential error propagation issue (shown in Figure~\ref{fig:consistency}).
Specifically, multi-task self-consistency contains the sentence-level self-consistency for intent detection and token-level self-consistency for slot filling.
\subsubsection{Sentence-level Self-consistency for Intent detection}
As shown in Figure~\ref{fig:consistency}, for the sentence-level intent detection task, we employ a direct voting mechanism to produce a unanimous inference, which can be formalized as follows:

\begin{small}
	\begin{equation}
		\hat{\mathcal{A}}^I =\mathop{argmax}_{\mathcal{A}^I \in {N^I}} \sum_{r=1}^{|R|} \mathds{1} \left(\mathcal{A}^I=\mathcal{Y}^I_r\right),
	\end{equation}
\end{small}
where $|R|$ is the number of consistency routes and $N^I$ represents the set of $\mathcal{Y}^I_r$ predicted by all routes.
Additionally, $\mathds{1} \left(X\right)$ represents a binary function that returns 0 if $X$ is False and 1 if $X$ is True. As shown in the example in Figure~\ref{fig:consistency}, among the four prediction results, "\texttt{SearchMovie}" appears the most frequently. Therefore, the final prediction result after self-consistency is "\texttt{SearchMovie}".

\subsubsection{Token-level Self-consistency for Slot Filling}
For the token-level slot filling task, the complexity lies out in two dimensions: the first dimension is the determination of the position of the slot-value, and the second dimension is the determination of the slot-type for each selected slot-value.

Our strategy first employs voting to establish the positon of slot value $\mathcal{A}^{value}_i$, which can be formally presented as follows:

\begin{small}
	\begin{align}
		& \hat{\mathcal{A}}^{value}_i \in \{\mathop{count}(\mathcal{Y}^{value}_{ir}) \geq \frac{|R|}{2}, \forall  1\leq r \leq |R| \},
	\end{align}
\end{small}
where $\mathcal{Y}^{value}_{ir}$ refers to the potential predicted slot value position and $|R|$ refers to the number of consistency routes. When more than half of the routes predict the same slot value, we consider it as a final prediction result after self-consistency.
As shown in Figure~\ref{fig:consistency}, for the sentence "\textit{Find the movie The Ghost}", more than half of the predictions identify "\texttt{The Ghost}" as a slot span.
So we choose "\texttt{The Ghost}" as a final slot value.

In step 2, we identify the slot type $\mathcal{A}^{type}_i$ for each voted $\mathcal{A}^{value}_i$ . This can be formally articulated as follows:

\begin{small}
	\begin{align}
		&\hat{\mathcal{A}}^{type}_i \!\!=\!\mathop{argmax}_{\mathcal{A}^{type}_{i} \in {N^S}} \sum_{r=1}^{|R|}  \mathds{1}\! \left(\!\mathcal{A}^{type}_{i}\!\!=\!\mathcal{Y}^{type}_{ir} | \hat{\mathcal{A}}^{value}_i \right),
	\end{align}
\end{small}
where $\mathcal{Y}^{type}_{ir}$ refers to the potential predicted slot type for each route predicting $\mathcal{A}^{value}_i$.
We select the slot type that appears most frequently.
As shown in Figure~\ref{fig:consistency}, we ensemble the slot type predictions of "\texttt{The Ghost}" to obtain its final slot type "\texttt{movie\_name}".

\begin{table*}
	\small
	\centering
	\begin{adjustbox}{width=0.78\textwidth}
		
		\begin{tabular}{lllll}
			\toprule
			\multirow{2}{*}{\textbf{Model}} & &\multicolumn{3}{c}{SNIPS}\\		
			\cmidrule(r){3-5}
			& & \multicolumn{1}{c}{Sentence Acc. (\%)} & \multicolumn{1}{c}{Intent Acc. (\%)}      & \multicolumn{1}{c}{Slot F1 (\%)} \\
			\midrule

			\multirow{6}{*}{\textsc{AgentLM-7B}~\cite{zeng2023agenttuning}}& \textit{Vanilla Prompting} & 4.43 & 84.00 & 26.22 \\
			&\textit{~~+Self-consistency} & {4.86}  & {84.29}  & {31.89}\\
			&\textit{CoT} & {4.43}  & {83.14}  & {24.84}\\
			&\textit{Self-Refine} & {5.14}  & {68.14}  & {25.75}\\
			&\textit{Plan-and-Solve} & {3.71}  & {57.86}  & {20.05}\\
			& \textbf{\textit{CroPrompt}}  & 8.00 \textcolor{blue!90!black}{(+3.57)} & 83.14 \textcolor{red!80!black}{(-0.86)} & 28.19 \textcolor{blue!90!black}{(+1.97)} \\
			&\textit{~~+MT-Self-consistency} & \textbf{8.29 \textcolor{blue!90!black}{(+3.86)}} & \textbf{85.14 \textcolor{blue!90!black}{(+1.14)}} & \textbf{34.17 \textcolor{blue!90!black}{(+7.95)}} \\
			\midrule
			
			\multirow{6}{*}{\textsc{Llama3-8B}~\cite{llama3}}& \textit{Vanilla Prompting} & 19.00 & 80.57 & 56.03 \\
			&\textit{~~+Self-consistency} & {19.00}  & {80.71}  & {56.45}\\
			&\textit{CoT} & {20.57}  & {82.43}  & {57.80}\\
			&\textit{Self-Refine} & {16.00}  & {80.29}  & {56.64}\\
			&\textit{Plan-and-Solve} & {21.71}  & {80.57}  & {58.84}\\
			& \textbf{\textit{CroPrompt}}  & 35.00 \textcolor{blue!90!black}{(+16.00)}  & 93.43 \textcolor{blue!90!black}{(+12.86)} & 66.52 \textcolor{blue!90!black}{(+10.49)}\\
			&\textit{~~+MT-Self-consistency} & \textbf{36.71 \textcolor{blue!90!black}{(+17.71)}} & \textbf{93.43 \textcolor{blue!90!black}{(+12.86)}} & \textbf{67.66 \textcolor{blue!90!black}{(+11.63)}} \\
			\midrule

			\multirow{6}{*}{\textsc{GPT-3.5-Turbo}~\cite{openai2022gpt35}}& \textit{Vanilla Prompting} & 35.75  & 95.51   & 69.02 \\
			&\textit{~~+Self-consistency} & {35.89}  & {95.51}  & {69.01}\\
			&\textit{CoT} & {38.92}  & {97.97}  & {67.92}\\
			&\textit{Self-Refine} & {33.86}  & {94.79}  & {64.88}\\
			&\textit{Plan-and-Solve} & {39.51}  & \textbf{98.12}  & {68.54}\\
			&\textbf{\textit{CroPrompt}}  & {40.96} \textcolor{blue!90!black}{(+5.21)} & {95.66} \textcolor{blue!90!black}{(+0.15)} & {71.79} \textcolor{blue!90!black}{(+2.77)} \\
			&\textit{~~+MT-Self-consistency} & \textbf{43.56 \textcolor{blue!90!black}{(+7.81)}} & {95.66 \textcolor{blue!90!black}{(+0.15)}}  & \textbf{75.32 \textcolor{blue!90!black}{(+6.30)}} \\
			\midrule
			\multirow{6}{*}{\textsc{GPT-4}~\cite{openai2023gpt4}}& \textit{Vanilla Prompting}  & 53.84 &  {99.28} & 80.39\\
			&\textit{~~+Self-consistency} & {52.97}  & \textbf{99.42}  & {79.95}\\
			&\textit{CoT} & {51.95}  & {98.84}  & {78.64}\\
			&\textit{Self-Refine} & {52.10}  & {98.99}  & {78.94}\\
			&\textit{Plan-and-Solve} & {54.56}  & {99.27}  & {79.49}\\
			&\textbf{\textit{CroPrompt}} & {67.00} \textcolor{blue!90!black}{(+13.16)}  & {98.84} \textcolor{red!80!black}{(-0.44)}  &{84.66} \textcolor{blue!90!black}{(+4.27)}  \\
			&\textit{~~+MT-Self-consistency} & \textbf{68.16 \textcolor{blue!90!black}{(+14.32)}} & 98.70 \textcolor{red!80!black}{(-0.58)} & \textbf{86.56 \textcolor{blue!90!black}{(+6.17)}} \\
			\bottomrule
		\end{tabular}
	\end{adjustbox}
	
	\caption{Main Results. For \textit{Vanilla Prompting}, we follow \citet{pan2023preliminary} to directly utilize a simple single-round prompting method for SLU.  Performance gains/drops compared to \textit{Vanilla Prompting} are highlighted with \textcolor{blue!90!black}{blue} / \textcolor{red!80!black}{red}. The best results are illustrated in bold.} 
	\label{tab:main_results}
\end{table*}

\section{Experiments}

\subsection{Datasets and Baselines}
We evaluate slot filling and intent detection on the widely used benchmark SNIPS~\citep{coucke2018snips}.
Besides, To verify the generalization of our framework, we also employ CroPrompt on other dialogue-correlated tasks (detailed analysis can be found in Section$\S \ref{sec:downstream}$).

In this work, we evaluate CroPrompt with some representative backbones, including: \textit{AgentLM}~\citep{zeng2023agenttuning}, \textit{Llama-3-8B}~\citep{llama3}, \textit{GPT-3.5-Turbo}~\citep{openai2022gpt35} and \textit{GPT-4}~\citep{openai2023gpt4}.
In addition, we conduct experiments on the recent baselines:
\begin{itemize}
	\item \textit{Vanilla Prompting}~\citep{pan2023preliminary} is a straightforward prompt method that accomplishes both intent detection and slot filling tasks in a single conversation turn.
	\item \textit{CoT}~\citep{NEURIPS2022_8bb0d291} is widely used in reasoning tasks, improving model performance by generating the reasoning process.
	\item \textit{Self-Refine}~\citep{madaan2023selfrefine} improves performance by modifying incorrect answers through the model's self-feedback.
	\item \textit{Plan-and-Solve}~\citep{wang-etal-2023-plan} uses Plan and Solve prompting method to improve
	the quality of generated reasoning steps.
	
\end{itemize}

\subsection{Evaluation Metric}
For slot filling and intent detection tasks, we follow \citet{goo-etal-2018-slot} and \citet{qin-etal-2020-dynamic} to measure the performance of intent detection and slot filling by intent accuracy and slot F1 score, respectively. Furthermore, sentence accuracy is also adopted to evaluate the accuracy of sentences that are predicted correctly for both intent and slot.

\subsection{Main Result}

The main results are shown in \tablename~\ref{tab:main_results}.
Our observations are as follows:

\begin{itemize}
	\item 
	\textbf{\textit{CroPrompt Beats Previous Baseline.}}
	For all four of the large language models (LLM), namely AgentLM, Llama3, GPT-3.5-Turbo, and GPT4, the performance of \textit{CroPrompt} surpasses that of the \textit{Vanilla Prompting} model by a significant margin. Specifically, 
	with the advanced GPT4 model, we achieve a substantial 13.16\% elevation in Sentence Acc. and an impressive 4.27\% improvement in Slot F1 score. 
	This indicates that \textit{CroPrompt} effectively captures the information exchange  between intent detection and slot filling, which leads to higher performance.
	We can also observe that \textit{CroPrompt + MT-Self-consistency} outperforms \textit{Vanilla Prompting + Self-consistency} by a significant extent, which demonstrates the effectiveness of \textit{CroPrompt}.

	\item 	\textbf{\textit{Self-consistency Boosts Performance.}}
	When comparing \textit{CroPrompt + MT-Self-consistency} to \textit{CroPrompt}, we observe improvements of 0.29\%, 1.71\%, 2.60\%, and 1.16\% in Sentence Acc. for the four respective models. 
	This indicates that our \textit{Multi-Task Self-consistency Prompting} helps mitigate the issue of error propagation, leading to an improvement in overall performance.

	\item \textbf{\textit{Better LLM, Better Performance.}}
	Lastly, we observe that \textit{GPT-4} with \textit{CroPrompt + Self-consistency} achieves the best performance with 86.56\% Slot F1 and 68.16\% Sentence Acc.
	Such observation demonstrates that a stronger LLM can attain better performance.
	
\end{itemize}

\subsection{Analysis}

To achieve a deeper comprehension of 
our framework, we conduct extensive analysis with \textit{GPT-3.5-Turbo} to answer the following question:
(1) Does the information transfer in LLM between tasks enhance performance?
(2) Is our \textit{CroPrompt} robust to the task order?
(3) What is the performance with gold intent information?
(4) What are the impacts of different methods of self-consistency?
(5) Can \textit{CroPrompt} method generalize to other tasks?
(6) Does \textit{CroPrompt} require more token costs?
(7) Is \textit{CroPrompt} applicable for domain adaptation SFT?
(8) Why \textit{CroPrompt} works?

The specific prompt texts for \textit{CroPrompt} and \textit{Vanilla Prompting} are detailed in the Appendix~\ref{sec:prompts}.

\subsubsection{Answer1: Information Exchange across Tasks Boost Performance}
\begin{table}[t]
	\centering
	\small
	\begin{adjustbox}{width=0.45\textwidth}
		\begin{tabular}{lccc}
			\toprule
			\multirow{2}{*}{\textbf{Method}} & \multicolumn{3}{c}{SNIPS}\\
			\cmidrule(r){2-4}
			& Sentence Acc & Intent Acc      & Slot F1       \\
			\midrule
			\textit{No-Interaction} & 35.46 & 95.66 & 69.30 \\
			\textit{CroPrompt}  & \textbf{40.96} & 95.66  & \textbf{71.79}   \\
			\bottomrule
		\end{tabular}
	\end{adjustbox}
	\caption{\textit{CroPrompt} vs. \textit{No-Interaction}. \textit{No-Interaction} accomplishes intent detection and slot filling in two separate sessions without task interaction.
	}
	\label{tab:interaction}
\end{table}

To verify whether the information exchange within \textit{CroPrompt} contributes to performance improvement, we conduct an experiment by splitting the intent detection and slot filling into different dialogue sessions, addressing these tasks independently without any information sharing.
We refer to this experiment as the \textit{No-Interaction} approach.

The results are illustrated in \tablename~\ref{tab:interaction}.
We observe that the metrics of \textit{No-Interaction} on Slot F1 and Sentence Acc drops by 2.49\% and 5.50\% compared to \textit{CroPrompt}, which demonstrates that explicitly leveraging task interaction can enhance the performance of related tasks.

\subsubsection{Answer2: CroPrompt is Robust to the task interaction order}
\begin{table}[t]
	\centering
	\small
	\begin{adjustbox}{width=0.48\textwidth}
		\begin{tabular}{lccc}
			\toprule
			\multirow{2}{*}{\textbf{Method}} & \multicolumn{3}{c}{SNIPS}\\
			\cmidrule(r){2-4}
			& Sentence Acc & Intent Acc      & Slot F1      \\
			\midrule
			\textit{Vanilla Prompting}  & 35.75 & 95.51   & 69.01 \\
			\midrule
			\textit{CroPrompt (Intent-Slot)}  & \textbf{40.96}  & 95.66  & \textbf{71.79} \\
			\textit{CroPrompt (Slot-Intent)}   & 36.03 & \textbf{96.53}  & 69.30\\
			\bottomrule
		\end{tabular}
	\end{adjustbox}
	\caption{Performance of different task orders for \textit{CroPrompt}. \textit{Intent-Slot} is consistent with \textit{CroPrompt} in Table~\ref{tab:main_results}. \textit{Slot-Intent} first completes the slot filling task, followed by completing the intent detection task.}
	\label{tab:order}
\end{table}
This section explores the robustness of \textit{CroPrompt} by testing the influence of task interaction order. Specifically, we investigate two task interaction manners: (1) one that predicts the intent first and then predicts the slot in the second turn, referred to as the \textit{CroPrompt (Intent-Slot)} mode, and (2) another that predicts the slot first and then predicts the intent in the second round, referred to as the \textit{CroPrompt (Slot-Intent)}.

The results are presented in Table~\ref{tab:order}. We have the following interesting discoveries:
(1) Both the \textit{CroPrompt (Intent-Slot)} and \textit{CroPrompt (Slot-Intent)} methods outperform  \textit{Vanilla prompting}, which indicates the robustness of CroPrompt;
(2) For the \textit{CroPrompt (Intent-Slot)}, we achieve the highest performance in slot filling, whereas for the \textit{CroPrompt (Slot-Intent)}, we obtain the highest performance in intent detection. This further validates that explicitly providing prior knowledge from the previous task can boost the performance of the subsequent related task.

\subsubsection{Answer3: Gold Intent can Further Boost Performance}

\begin{table}[t]
	
	\centering
	\small
	\begin{adjustbox}{width=0.42\textwidth}
		\begin{tabular}{lccc}
			
			\toprule
			\multirow{2}{*}{\textbf{Method}} & \multicolumn{3}{c}{SNIPS}\\
			\cmidrule(r){2-4}
			& Sentence Acc & Intent Acc      & Slot F1       \\
			\midrule
			\textit{CroPrompt}  & {40.96} & 95.66  & {71.79}  \\
			\textit{Gold Intent} & \textbf{44.14} & \textbf{100.00} & \textbf{72.48} \\
			\bottomrule
		\end{tabular}
	\end{adjustbox}
	\caption{Performance with Gold Intent. \textit{Gold Intent} treats the intent gold label as the output result.}
	\label{tab:gold-intent}
\end{table}
To demonstrate the effectiveness of \textit{CroPrompt}, we conduct an experiment involving the utilization of gold intent detection results.

Results are displayed in Table~\ref{tab:gold-intent}. We observe that when utilizing gold intent information, an improvement of 0.69\% in Slot F1 and 3.18\% in Sentence Acc. is achieved. This observation suggests that better intent detection performance can attain better improvement in \textit{CroPrompt}, which further verifies that information exchange is crucial for SLU.

\subsubsection{Answer4: Investigation of Different Self-Consistency Methods}

\begin{figure}[t]
	\centering
	\includegraphics[width=.45\textwidth]{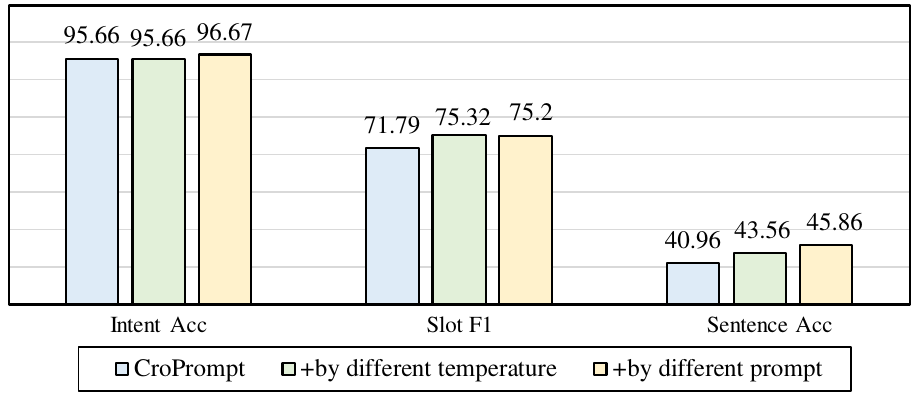}
	\caption{Results of \textit{CroPrompt} with different multi-task self-consistency methods. \textit{By different temperature} utilizes the outputs of \textit{CroPrompt} with different temperatures. \textit{By different prompt} utilizes the outputs of three different prompt methods.}
	\label{fig:different-consistency}
\end{figure}
In this experiment, we investigate two different consistency approaches, named \textit{MT-Self-Consistency by different temperature} and \textit{by different prompt}. 
For \textit{by different temperature}, we sample the ChatGPT's responses with temperature in [0.1, 0.8, 1.0], which are all conducted with the best \textit{CroPrompt} setting.
For \textit{by different prompt}, we use three different prompt including \textit{Vanilla Prompting}, \textit{CroPrompt (Intent-Slot)} and \textit{CroPrompt (Slot-Intent)}. The three different prompts represent distinct information interaction: no-interaction, intent to slot and slot to intent.

The results are shown in Figure~\ref{fig:different-consistency}.
In comparison to the standard \textit{CroPrompt}, we observe an improvement of 1.01\% in intent Acc. in the \textit{MT-self-consistency by different prompt}. Moreover, the \textit{MT-self-consistency by different prompt} approach yields a significant increase of 3.41\% in slot F1 score. 
This suggests that the self-consistency exhibits a significant performance improvement.
In addition, we observe that the \textit{by different prompt} outperforms \textit{by different temperature}  by 1.01\% and 2.30\%  for intent Acc. and sentence Acc., respectively. 
We credit it to the cause that the \textit{by different prompts} incorporates three different modes of information exchange, which makes it achieve the best performance.

\subsubsection{Answer5: Generalizing to Other Downstream Task}\label{sec:downstream}

\begin{figure}[t]
	\centering
	\includegraphics[width=.45\textwidth]{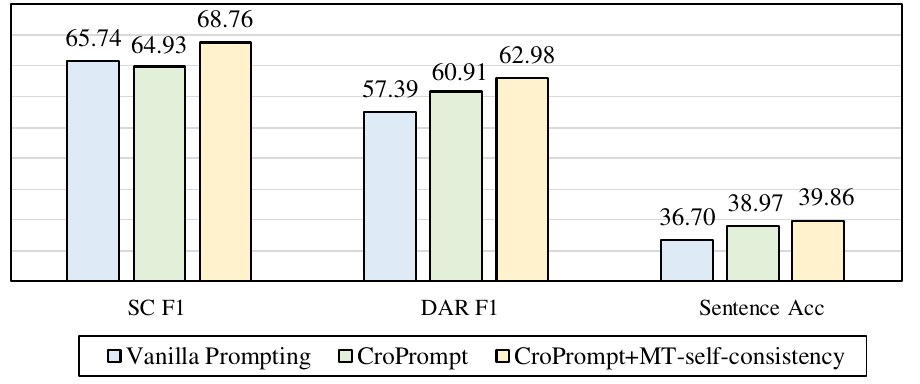}
	\caption{Results of \textit{CroPrompt} for Join SC and DAR on Mastodon.}
	\label{fig:mastodon}
\end{figure}

An engaging question that surfaces is whether \textit{CroPrompt} can generalize to other tasks. To answer this question, we apply \textit{CroPrompt} to dialogue act recognition and dialogue sentiment classification that are two correlated sentence-level classification tasks in dialogue understanding. The former is used for detecting the explicit intent of the user, while the latter can identify implicit intentions. Specifically, we first employ the \textit{preliminary task solution prompting} for dialogue sentiment classification and then utilize the \textit{follow-up task solution prompting} for dialogue act recognition.

The results are presented in Figure~\ref{fig:mastodon}.
We can see that \textit{CroPrompt} attains superior performance in terms of Sentence Accuracy when compared to \textit{Vanilla Prompting}, showing an improvement of 2.27\%. Moreover, when employing the multi-task self-consistency approach, we observe additional enhancements in performance.  These results are consistent with the observations in SLU, which further demonstrates the effectiveness of \textit{CroPrompt}.

\subsubsection{Answer6: CroPrompt can Reduce API Token Cost}

\begin{table}[t]
	\centering
	\begin{adjustbox}{width=0.35\textwidth}
		\begin{tabular}{lc}
			\toprule
			\textbf{Method} & Context Average Length\\
			\midrule
			\textit{Vanilla Prompting} & 619.42  \\
			\textit{CroPrompt} & 299.04 \\
			\bottomrule
		\end{tabular}
	\end{adjustbox}
	\caption{The context length of \textit{Vanilla Prompting} and proposed \textit{CroPrompt}.}
	\label{tab:context-length-comparison}
\end{table}

To explore whether our proposed \textit{CroPrompt} can reduce the context length by providing intent-constrained slot descriptions, as mentioned in Section~\ref{section:mmprompt}. We calculate the context length of \textit{Vanilla Prompting} and \textit{CroPrompt}.

Table~\ref{tab:context-length-comparison} shows the context average length of the \textit{Vanilla Prompting} and \textit{CroPrompt}. 
We can observe that on the SNIPS dataset, the  \textit{Vanilla Prompting}  has a context length that is 2.07 times longer than \textit{CroPrompt}. We attribute it to the fact that since \textit{CroPrompt} can obtain the intent information in the first round, we only need to provide a subset of the slot description given the predicted intent to the model, resulting in decreased latency and API token cost.

\subsubsection{Answer7: CroPrompt exhibits stronger generalization capabilities in SFT.}
\label{sec:sft}
\begin{table}[t]
	\centering
	\small
	\begin{adjustbox}{width=0.49\textwidth}
		\begin{tabular}{llccc}
			\toprule
			\multirow{2}{*}{\textbf{Model}} & & \multicolumn{3}{c}{SNIPS}\\
			\cmidrule(r){3-5}
			&  & Sentence Acc &Intent Acc      & Slot F1     \\
			\midrule
			\multirow{2}{*}{{Llama3-8B}} & \textit{Vanilla Prompting} & {19.00}  & {80.71}  & {56.45}\\
			& \textit{CroPrompt} & 35.00  & 93.43  & 66.52 \\
			
			\midrule
			\multirow{2}{*}{{Llama3-8B-SFT}} & \textit{Vanilla Prompting} & 36.00 & 89.00 & 66.49\\
			& \textit{CroPrompt} & \textbf{51.86} & \textbf{96.29} & \textbf{74.54}\\
			
			\bottomrule
		\end{tabular}
	\end{adjustbox}
	\caption{Results of conducting supervied finetuning (SFT) on \textit{Llama3-8B} using ATIS dataset to verify the domain generalization capability.
	}
	\label{tab:sft}
\end{table}

We explore whether \textit{CroPrompt} can help open-source models further improve model performance through supervised fine-tuning (SFT). We finetune \textit{Llama-3-8b-chat} on ATIS \citep{hemphill-etal-1990-atis} dataset to verify the domain generalization capabilities of \textit{Vanilla Prompting} and \textit{CroPrompt}. The prompt structure used for inference matches the format employed during training.
We utilze LoRA\citep{hu2021lora} to finetune \textit{Llama3-8B-Chat} for the two prompts. We set the lora-rarget to \texttt{q\_proj} and \texttt{v\_proj}. For both methods we finetune for 2 epochs with batch\_size = 32, learning\_rate = 0.0001 and lora\_rank = 32. We use a consine lr scheduler with 0.1 warmup steps.

As shown in Table~\ref{tab:sft}, both methods have led to improvements in the domain transfer SFT setting. SFT with \textit{CroPrompt} achieved the best performance in both intent detection and slot filling, with an improvement of 7.29\% and 8.05\% compared to SFT with \textit{Vanilla Prompting}. 
On one hand, the two-stage processing of \textit{CroPrompt} makes it easier for the model to acquire the capabilities of both tasks. On the other hand, \textit{Vanilla Prompting} has only one type of prompt format for all data, whereas \textit{CroPrompt} uses different slot descriptions for different intents, and the diversity of prompts can lead to better generalization capabilities.

\subsubsection{Answer8: Qualitative analysis}
\label{sec:qualitative}
\begin{figure*}[t]
	\centering
	\includegraphics[width=0.85\textwidth]{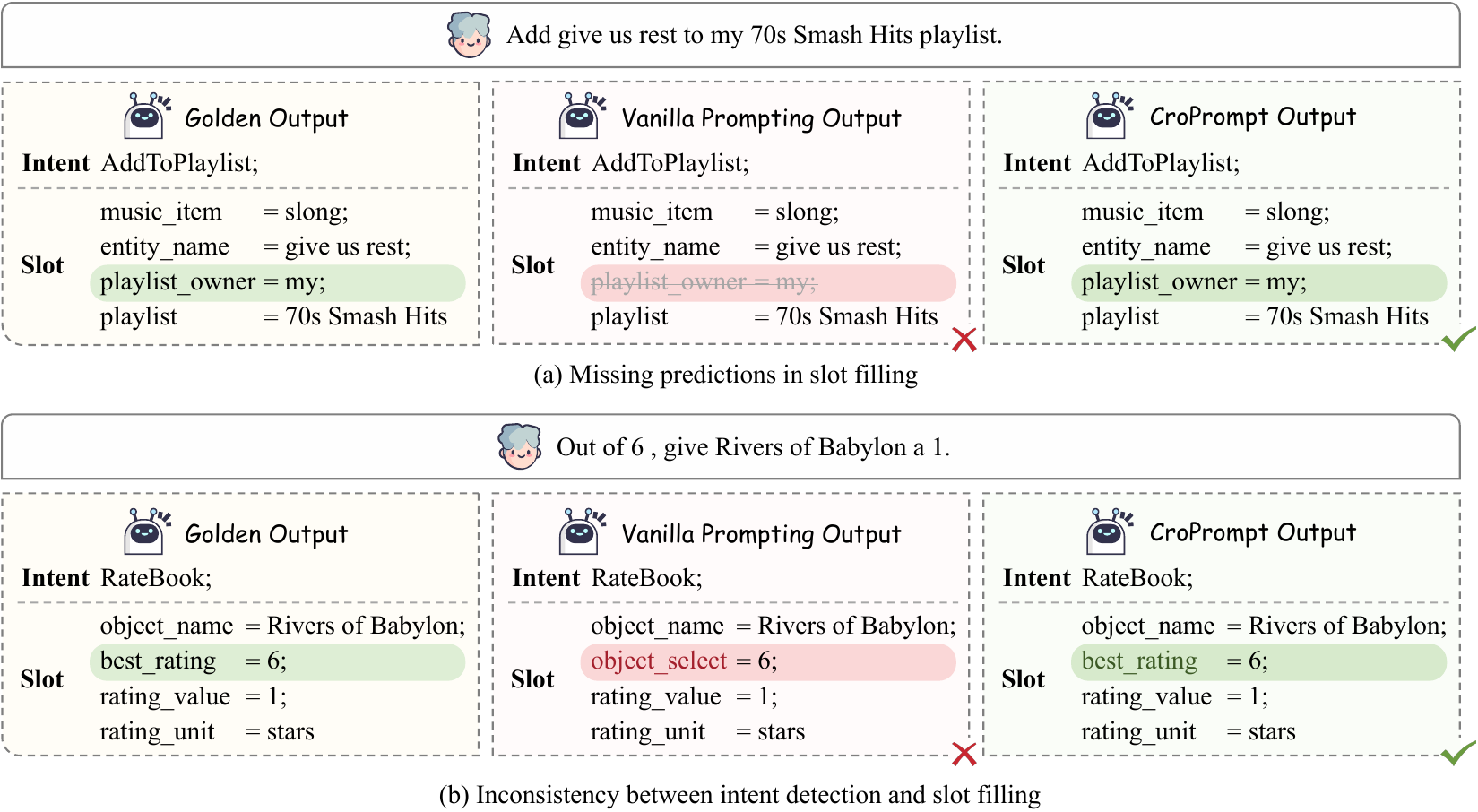}
	\caption{Case Study of \textit{CroPrompt} and \textit{Vanilla Prompting}. 
	}
	\label{fig:case}
\end{figure*}

To gain a clearer insight into the functioning of \textit{CroPrompt}, we offer a detailed case study generated by \textit{CroPrompt} and \textit{Vanilla Prompting}.
Compared to \textit{CroPrompt}, the \textit{Vanilla Prompting} approach has two main drawbacks, which are shown in Figure~\ref{fig:case}:

(1) \textbf{Missing predictions in slot filling}: As shown in Figure~\ref{fig:case} (a), the \textit{Vanilla Prompting} frequently exhibits slot omissions (\textit{slot "playlist\_owner" is omitted}). This is because that all tasks are expected to be completed within a single dialogue turn, which makes it hard for ChatGPT to obtain the correct output for slot filling. Additionally, the \textit{Vanilla Prompting} lacks information interaction between the two tasks, making it challenging to achieve effective slot filling.

(2) \textbf{Inconsistency between intent detection and slot filling}: As demonstrated in Figure~\ref{fig:case} (b), the \textit{Vanilla Prompting} generates slots that are inconsistent with the predicted intent (\textit{slot "object\_slelect" is contradicted with intent "RateBook"}).
In contrast, \textit{CroPrompt} generates the correct intent and slot pairs. We attribute it to the fact that the explicit intent information exchange and injection can effectively mitigate the issue of intent-slot inconsistencies.

\section{Related Work}

Spoken language understanding (SLU) 
is a core task in task-oriented dialogue system, which typically consists of two tasks: intent detection and slot filling~\citep{ijcai2021p622}.
For the two related subtasks, some work models intent detection and slot filling through parameter-sharing implicit interaction.
\citet{DBLP:conf/ijcai/ZhangW16a} leverage a shared
RNNs to model the correlation between intent detection and slot filling. \citet{DBLP:journals/corr/LiuL16d} introduce a parameter-shared BiRNN with attention mechanism for joint modeling.
\citet{DBLP:conf/interspeech/Hakkani-TurTCCG16} introduce Joint Seq with a shared RNN-LSTM for intent and slots.
Other work enhances the modeling of intent detection and slot filling through explicit interaction.
A series of studies leverage the information of intent detection to guide slot filling task using explicit network unidirectional interaction architecture~\citep{goo-etal-2018-slot,li-etal-2018-self,qin-etal-2019-stack,9413657}.
After that, various work focus on bidirectional interaction modeling has emerged~\citep{wang-etal-2018-bi,DBLP:conf/acl/ENCS19,zhang-etal-2019-joint,liu-etal-2019-cm,DBLP:conf/aaai/ZhangMZYW20}.
\citet{9414110} further propose the Co-Interactive framework, which displays bidirectional interaction between intention and slot information with Transformer structure. 
Meanwhile, some work explores profile-SLU, which aims to mitigate ambiguity issues among users in real SLU scenarios by leveraging user profile information~\citep{xu2022text,teng2024pro}.
Nevertheless, the above approaches still rely on a large amount of annotated data for training, which is hard to collect.
As LLMs have demonstrated strong performance across various NLP tasks \citep{brown2020language,liu2023pre}, certain studies have initiated investigations into SLU tasks using LLMs.
\citet{pan2023preliminary} propose a unified prompt to jointly solve intent detection and slot filling tasks.
Expanding upon this foundation, \citet{DBLP:journals/corr/abs-2305-13512} introduce a similar unified prompt and analyze it under more settings such as the smaller LLMs and contextual examples. 

In contrast to their approach, this work explores a cross-task prompting approach for SLU, which has the advantage of explicitly incorporating the inter-task information exchange. Recently, \citet{zhu-etal-2024-zero-shot} preliminarily explore task interaction of SLU tasks on ChatGPT. Different from their work, we explore more closed-source and open-source models. In addition, we further introduce a multi-task self-consistency prompting to mitigate the error propagation.

\section{Conclusion}
In this work, we explore the cross-task prompting for zero-shot spoken language understanding and introduce a novel Cross-task Interactive Prompting (CroPrompt) to this end, achieving interactively leveraging the information exchange across related tasks.
In addition, we further present a multi-task self-consistency prompting to mitigate the error propagation issue.
Experimental results reveal that CroPrompt outperforms previous methods and multi-task self-consistency prompting can further consistently gain improvement.
\section*{Limitations}
This work represents a pioneering investigation of cross-task prompting for zero-shot spoken language understanding. However, due to the autoregressive decoding nature of LLMs, our CroPrompt mainly focuses on the unidirectional information exchange from the preliminary task to the follow-up task. In the future, we can explore achieving bidirectional information exchange within a single session.

Besides, there is a significant performance gap between open-source models with smaller parameter sizes (e.g. Llama3-8B) and closed-source LLMs (e.g. GPT-4). We observe that smaller LLMs show inadequacies in output format adherence and task comprehension. Open-source models with fewer parameters are more resource-efficient and cost-effective. Therefore, enhancing the zero-shot SLU abilities of small LLMs remains to be explored in the future.

\bibliographystyle{named}
\bibliography{ijcai24,aaai24,custom,anthology}

\clearpage
\appendix
\section{Appendix}

\subsection{Prompts}
\label{sec:prompts}
We follow the prompt text of baseline \textit{Vanilla Prompting} and only separate the intent and slot parts into two rounds without more modification. The full prompt of \textit{Vanilla Prompting} and \textit{CroPrompt} are displayed in Table~\ref{tab:vallina_prompt} and Table~\ref{tab:full_crompt}.

\begin{table*}[b]
	\centering
	\begin{adjustbox}{width=0.90\textwidth}
		\begin{taskbox}[]{Query}
			
			\TaskA{[Task Instruction]}

			You need to annotate some sentences I gave you in the following, which includes intent and slots.\\
			
			\LabelSet{[Task Label Constraint]}

			Given following sentences, first choose the intent of the sentences from the following intent list: [\texttt{AddToPlaylist; BookRestaurant;}...]. \\
			
			Then annotate given sentences with slots from following slot list, the description of each slot is given.\\
			\texttt{album}: Name of the album that user want to ...\\
			\texttt{artist}: Name of musical artist mentioned ...\\
			\texttt{best\_rating}: Max rating stars/points of ...\\
			\texttt{city}: Name of the city request by the user ...\\
			\texttt{condition\_description}: Weather condition queried ...\\
			\texttt{condition\_temperature}: Temperature condition when quering ...\\
			......\\
			\\
			
			\Regulation{[Task Regulation]}

			You need to output the annotations in the form of ``Intent=INTENT\_NAME; Slot1=VALUE1; Slot2= VALUE2; ..." \\			
			You must not output anything other than the annotations. \\
			You must not miss any possible slot-value pairs.\\
			
			\InputName{[Given Sentence]}

			Here is the the sentence:\\
			\texttt{put United Abominations onto my rare groove playlist}
			
			\tcbsubtitle{LLM response}
			Intent=\texttt{AddToPlaylist}; \texttt{entity-name}=``United Abominations"
			
		\end{taskbox}
	\end{adjustbox}
	\caption{Example of Vanilla Prompting.}
	\label{tab:vallina_prompt}
\end{table*}

\begin{table*}[h]
	\centering
	\begin{adjustbox}{width=0.90\textwidth}
		\begin{taskbox}[]{Intent Detection Query}
			
			\TaskA{[Intent Task Instruction]}

			Given following sentences, first choose the intent of the sentences from the following intent list: \\
			
			\LabelSet{[Intent Label Constraint]}

			[\texttt{AddToPlaylist; BookRestaurant;}...]. \\
			
			\Regulation{Intent Regulation}

			You need to output the intent annotations in the form of "Intent=INTENT\_NAME"
			You must not output anything other than the intent annotations.\\
			
			\InputName{[Given Sentence]}

			Here is the the sentence:\\
			\texttt{put United Abominations onto my rare groove playlist}
			
			\tcbsubtitle{LLM Intent detection response}
			Intent=\texttt{AddToPlaylist}

			\tcbsubtitle{Slot Filling Query}
			\Answer{[Intent detection Answer]}

			Now you have annotated the sentence as \texttt{AddToPlaylist} intent.\\

			\TaskB{[Slot Task Instruction]}

			Then you annotate given sentences with slots from following slot list, the description of each slot is given.\\
			
			\LabelConstraint{[Slot Label Constraint]}

			\texttt{music\_item}: The type of item that user want to ...\\
			\texttt{entity\_name}: Name of the song to be added into ...\\
			\texttt{artist}: Name of musical artist mentioned in the ...\\
			\texttt{playlist}: Name of the playlist e.g. Flow ...\\
			\texttt{playlist\_owner}: Owner of the playlist e.g. my, ...\\
			
			\Regulation{[Slot Regulation]}

			You must not miss any possible slot-value pairs. 
			
			You need to output the annotations in the form of ``Slot1=VALUE1;Slot2=VALUE2;...''.
			
			You must not output anything other than the slot annotations.\\
			
			\InputName{[Given Sentence]}

			Repeat, the sentence is:\\
			\texttt{put United Abominations onto my rare groove playlist}
			\tcbsubtitle{LLM Slot Filling response}
			\texttt{entity-name}="United Abominations"

		\end{taskbox}
	\end{adjustbox}
	\caption{Example of CroPrompt.}
	\label{tab:full_crompt}
\end{table*}

\end{document}